\documentclass{article}
\usepackage{spconf,amsmath,graphicx,hyperref}
\usepackage{cite}
\usepackage{amsmath,amssymb,amsfonts}
\usepackage{algorithmic}
\usepackage{graphicx}
\usepackage{marvosym} 
\usepackage{amssymb}
\usepackage{amsmath}
\usepackage{multirow}
\usepackage{booktabs}
\usepackage{float}
\usepackage{xcolor}
\usepackage{colortbl}
\usepackage{array}
\usepackage{enumitem}
\usepackage{textcomp}
\usepackage{xcolor}

\title{MIRG-RL: Multi-Image Reasoning and Grounding with \\ Reinforcement Learning}
\name{\textbf{Lihao Zheng} \qquad \textbf{Jiawei Chen} \qquad \textbf{Xintian Shen} \qquad \textbf{Hao Ma} \qquad \textbf{Tao Wei}\textsuperscript{\Letter}\thanks{$^{\textrm{\Letter}}$Corresponding author.}}
\address{
\textbf{Li Auto Inc.}    \\
\\
\textit{zhenglihao@lixiang.com} }

\begin{document}
\ninept
\maketitle
%
\begin{abstract}

Multi-image reasoning and grounding require understanding complex cross-image relationships at both object levels and image levels. Current Large Visual Language Models\,(LVLMs) face two critical challenges: the lack of cross-image reasoning capabilities and insufficient cross-image reference reward modeling. To address these issues, we propose a unified framework—Multi-Image Reasoning and Grounding with Reinforcement Learning\,(MIRG-RL).
Specifically, our two-stage training paradigm combines supervised fine-tuning with annotated trajectories and image-aware reinforcement learning optimization, progressively developing multi-image reasoning capabilities.
Furthermore, we innovatively propose a method for constructing the trajectory data, which integrates object-level and image-level annotation information, and use this method to generate a lightweight reasoning-enhanced dataset.
To effectively resolve cross-image ambiguities, we design an image-aware RL policy with dual reward functions for objects and images.
Experiments demonstrate that MIRG-RL achieves state-of-the-art\,(SOTA) performance in multi-image grounding benchmarks, attaining 64.82\% on cross-image reasoning tasks—exceeding the previous best method by 1\%.
The code and dataset have been released at https://github.com/ZEUS2035/MIRG-RL.

\end{abstract}
\begin{keywords}
Multi-image reasoning and grounding, reinforcement learning, multi-modal large language model
\end{keywords}
%


\section{Introduction}
\label{sec:intro}


Multi-image reasoning and grounding are critical for multi-modal applications such as multi-view scene understanding and comparative visual question answering. These tasks require models to resolve fine-grained object references (e.g.,  "the man wearing a blue shirt in Image-1") and global image-level relations (e.g.,  "In Image-1, the object in Image-3 is located below the object in Image-4"). Compared to single-image scenes, this scene demands joint modeling of object-level and image-level contextual relationships, greatly increasing reasoning complexity. 

Previous visual grounding studies such as CLIP-VG\cite{xiao2023clip} and GLIP\cite{li2022grounded} relied on datasets like RefCOCO (e.g., \cite{mao2016generation, nagaraja2016modeling, yu2016modeling}) and were enhanced through visual-language pre-training frameworks, including CLIP \cite{radford2021learning}. These approaches mainly focused on phrase grounding and were restricted by template-based instructions. 

Subsequent research typically adopted multi-modal large language models (LLMs) such as Shikra~\cite{chen2023shikra}, GroundingGPT~\cite{li2024groundinggpt}, Ferret~\cite{you2023ferret}, and Griffon~\cite{zhan2024griffon}~\cite{zhan2024griffonv2} to explore how to perform single-image grounding in natural language dialogues. Others, such as LISA\_Grounding~\cite{lai2024lisa} and LLMSeg\_Grounding~\cite{wang2024llmseg}, explored grounding mechanisms through logical reasoning and segment-based guidance. 

Additionally, influenced by DeepSeek-R1~\cite{guo2025deepseek}, current reinforcement learning (RL) research is exploding. Visual-RFT~\cite{visual-rft} and VLM-R1~\cite{vlm-r1} extend RL to the single-image grounding domain. 
Nevertheless, these approaches have not considered a multi-image scenario and face substantial challenges in modeling complex cross-image relationships.

Recent studies such as Migician ~\cite{li2025migician} pioneer the first step in multi-image grounding research by proposing the large-scale MGrounding-630k dataset and employing a Supervised Fine-Tuning\,(SFT) strategy. However, such methods still suffer from a lack of demonstrated reasoning capabilities and fail to fully utilize multi-image information.

Given these limitations, existing methods face two key challenges: (i).\,Lack of reasoning across images: Current grounding models struggle to perform reasoning between multiple images. Most approaches either focus on single-image tasks or treat multi-image scenarios as independent single-image problems, failing to capture the complex inter-image dependencies essential for real-world applications. (ii).\,Insufficient reward modeling for cross-image references: 
Existing RL-based methods primarily design rewards within a single image, neglecting the need for explicit referential understanding and comprehension across multiple images at both the object and image levels. This limitation hinders models in resolving positional and comparative ambiguities during cross-image reasoning.
\begin{figure*}[t]
    \centering
    \includegraphics[width=\linewidth]{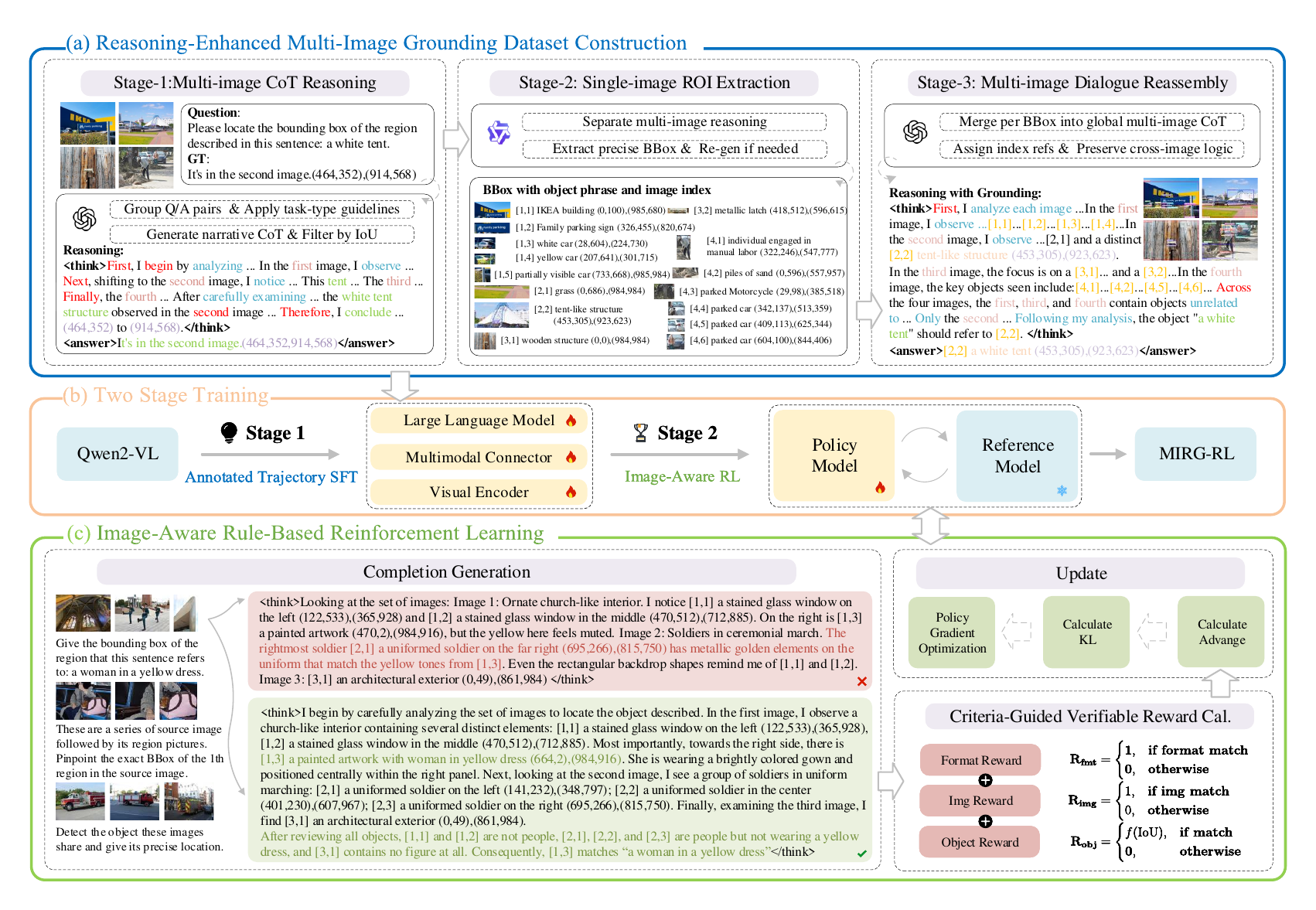}
    \caption{
    Illustration of the MIRG-RL pipeline for reasoning-enhanced multi-image grounding dataset construction and training.
    (a) Three-stage dataset construction: multi-image CoT reasoning, single-image ROI extraction, and multi-image dialogue reassembly.
    (b) Two-stage training framework: Stage~1 reasoning SFT based on annotated trajectories, and Stage~2 image-aware RL optimization.
    (c) Image-aware rule-based reinforcement learning process for completion generation, including criteria-guided verifiable reward calculation (format, image, and object rewards) and policy gradient optimization.
    }
    \label{method}
\end{figure*}

To address these challenges of multi-image reasoning and grounding, we propose a novel framework named \textbf{MIRG-RL}. This framework employs an innovative training paradigm and data construction method specifically designed to tackle the inherent complexity of multi-image scenarios. We evaluated our approach on multi-image grounded benchmark datasets featuring multiple tasks. MIRG-RL achieved state-of-the-art performance.

In summary, our contributions are as follows: 

(i).\,We propose a training framework that combines supervised fine-tuning of lightweight annotated reasoning trajectories with image-aware reinforcement learning optimization, enabling the model to learn and develop multi-image reasoning grounding capabilities. 

(ii).\,We innovatively propose a three-stage data construction method that employs multi-module visual reasoning and localization to generate novel inference trajectory data. This approach constructs a high-quality multi-image dataset featuring annotated inference trajectories, integrating both image-level and object-level reference information to advance comprehensive cross-image reasoning and localization capabilities. 

(iii).\,We pioneeringly introduce explicit image-level rewards through a dual reward function for objects and images, building upon rule-based RL to enable the model to more effectively resolve uncertainties in position and comparative relationships.

\section{Methods}
\label{sec:format}
\label{sec:method}
\subsection{Overview of MIRG-RL}

In this section, we present our method for multi-image reasoning and grounding. The task we address requires the model to perform precise object localization based on textual queries or visual reference prompts across multiple images. Formally, given a set of $N$ images $\mathcal{I} = \{I_1, I_2, ..., I_N\}$ and a natural language query $Q$, the model $\mathcal{M}$ needs to understand cross-image referential relationships and output precise grounding results $\mathcal{G} = \{g_1, g_2, ..., g_m\}$ for referenced objects, where each $g_i = (P_i, D_i, B_i)$ consists of position identifier, object description, and bounding box. Specifically, $P_i = (n, k)$ indicates the object is the $k$-th object in the $n$-th image, $D_i$ is the object description or caption, and $B_i = (x_1, y_1, x_2, y_2)$ represents the spatial coordinates.

As illustrated in Figure~\ref{method}(b), our training framework consists of two stages. First, we construct a lightweight high-quality dataset with Chain-of-Thought\,(CoT) annotations for supervised fine-tuning, enabling the model to learn structured reasoning trajectories. Second, we employ rule-based reinforcement learning approach to guide the model's reasoning process and enhance its comprehension capabilities.

\subsection{Reasoning-Enhanced Dataset Construction}
\label{sec:dataset}
Inspired by DeepSeek-R1~\cite{guo2025deepseek}, we initially explored training the model using pure reinforcement learning. 
However, the model lacks multi-image grounding capabilities and struggles to generate reasoning trajectories. 
Therefore, a well-constructed cold-start dataset is necessary to guide the model's learning of multi-image grounding and teach it how to reason effectively.

As shown in Figure~\ref{method} (a), we construct a dataset with chain-of-thought annotations through a three-stage pipeline:

\textbf{Stage 1: CoT Generation.} We provide GPT-4o~\cite{hurst2024gpt} with questions, bounding box, and task-specific prompts to generate reasoning processes following the format: \texttt{\small<think>thinking process</think><answer>answer</answer>}. The model analyzes the query content to identify relevant objects across images and produces step-by-step reasoning processes, explicitly capturing associative relationships between images.

\textbf{Stage 2: Image-Object Mapping.} We feed the generated data into Qwen2-VL-72B~\cite{wang2024qwen2} to establish mappings between reasoning trajectories and individual images. For each referenced object, the model generates fine-grained bounding boxes with image indices, ensuring accurate spatial localization for objects mentioned in the reasoning in process.

\textbf{Stage 3: Trajectory Reassembly.} We use GPT-4o to reorganize the reasoning trajectories, structuring each object with the format: \texttt{\small<bbox\_id>[N-M]</bbox\_id><|object\_ref}
\texttt{\small\_start|>[object]<|object\_ref\_end|><|box\_start|>}
\texttt{\small(x1,y1),(x2,y2)<|box\_end|>},
where N represents the image number and M represents the object number within that image. Subsequent references to the object use its bbox\_id. This structured format ensures consistent cross-image referencing throughout the reasoning process.

We use this dataset to perform supervised fine-tuning on the base model, resulting in our Stage-1 model. This model can generate final predictions through reasoning processes that explicitly include image indices and coordinate annotations.

\subsection{Image-Aware Reinforcement Learning}
\label{sec:rl}
In the second stage, we employ rule-based reinforcement learning to guide the model's reasoning process and enhance its understanding capabilities. Specifically, we adopt the Group Relative Policy Optimization\,(GRPO) algorithm. Given a query $q$, the GRPO algorithm samples $N$ responses $\{o_1, o_2, ..., o_N\}$ from the policy model $\pi_{\theta_{\text{old}}}$, then evaluates each response using reward function $R(q, o_i)$. 

Initially, we employ two basic reward functions for our task:

\textbf{Format Reward} ($r_{\text{fmt}}$): Ensures the response follows the required output format with proper think-answer structure, returning 1 for a valid format and 0 otherwise.

\textbf{Object Reward} ($r_{\text{obj}}$): Measures spatial grounding accuracy using IoU between predicted and ground-truth bounding boxes.
\begin{equation}
r_{\text{obj}} = \text{IoU}(B_{\text{pred}}, B_{\text{GT}})
\end{equation}

These rewards enable the model to improve its multi-image grounding capabilities through reinforcement learning. Building on insights from our dataset construction exploration, we introduce an additional reward function.

\textbf{Image Reward} ($r_{\text{img}}$): Validates correct image index assignment for each predicted object:

\begin{equation}
r_{\text{img}} = \begin{cases}
1, & \text{if } \pi_1(P_{\text{pred}}) = \pi_1(P_{\text{GT}}) \\
0, & \text{otherwise}
\end{cases}
\end{equation}
where $\pi_1(\cdot)$ represents the projection function that returns the first component from the position identifier.

The total reward for each response is computed as:
\begin{equation}
r =  r_{\text{fmt}} +  r_{\text{img}} + r_{\text{obj}}
\end{equation}

To determine relative response quality, GRPO normalizes rewards by computing advantages:
\begin{equation}
A_i = \frac{r_i - \text{mean}(\{r_1, r_2, ..., r_N\})}{\text{std}(\{r_1, r_2, ..., r_N\})}
\end{equation}

The policy is updated to maximize:
\begin{equation}
\begin{aligned}
\mathcal{J}_{\text{GRPO}}(\theta) = \mathbb{E}_{q \sim P(Q), \{o_i\}_{i=1}^N \sim \pi_{\theta_{\text{old}}}(O|q)} \\
\left[ \frac{1}{N}\sum_{i=1}^N \frac{\pi_{\theta}(o_i|q)}{\pi_{\theta_{\text{old}}}(o_i|q)}A_i - \lambda \mathbb{D}_{\text{KL}}(\pi_{\theta}||\pi_{\text{ref}}) \right]
\end{aligned}
\end{equation}

Through Stage-2 reinforcement learning, the model learns to select correct reasoning chains from multiple sampled responses, thereby enhancing its reasoning and grounding capabilities for multi-image scenarios.

\begin{table*}[t]
\setlength{\tabcolsep}{6pt}
\centering
\caption{Performance comparison of different models on MIG-Bench~\cite{li2025migician}. OT, MV, GG, and Co-Re respectively mean object tracking, multi-view grounding, group grounding, and correspondence. Our MIRG-RL achieves the best results across most tasks. The best results are shown in \textbf{bold}.}
\vspace{6pt}
\resizebox{\linewidth}{!}{%
\begin{tabular}{l|ccc|ccccccc|c}
    \toprule
    \multirow{3}{*}{\textbf{Models}} & \multicolumn{3}{c|}{\textbf{Spontaneous Grounding}} & \multicolumn{7}{c|}{\textbf{Referential Grounding}} & \multirow{3}{*}{\textbf{AVE}} \\ 
    \cmidrule(lr){2-4} \cmidrule(lr){5-11}
    & \multicolumn{2}{c|}{\textbf{Difference}} & \multirow{1}{*}{\textbf{Similarity}} & \multicolumn{4}{c|}{\textbf{Visual Reference}} & \multirow{1}{*}{\textbf{Textual}} & \multicolumn{2}{c|}{\textbf{Visual+Textual}} & \\ 
    \cmidrule(lr){2-3} \cmidrule(lr){4-4} \cmidrule(lr){5-8} \cmidrule(lr){9-9} \cmidrule(lr){10-11}
    & \textbf{Static} & \textbf{Robust} & \textbf{Common} & \textbf{OT} & \textbf{MV} & \textbf{Region} & \textbf{Refer} & \textbf{GG} & \textbf{Reason} & \textbf{Co-Re} & \\
    \midrule

\rowcolor{gray!10} \multicolumn{12}{c}{\textbf{70B-Scale MLLMs}}\\
\hline
LLaVA-OV-72B ~\cite{li2024llava1}  & 13.26 & 5.34 & 26.84 & 12.91 & 7.64 & 2.14 & 17.83 & 21.60 & 11.88 & 8.55 & 13.65 \\
InternVL2-76B ~\cite{chen2024expanding} & 15.91 & 10.64 & 36.40 & 30.73 & 20.83 & 5.74 & 46.46 & 41.28 & 32.67 & 26.50 & 26.72 \\
Qwen2-VL-72B ~\cite{wang2024qwen2} & 46.12 & 46.81 & 64.46 & 26.73 & 22.57 & 18.62 & 33.33 & 62.53 & 50.50 & 17.09 & 38.88 \\
\hline 
\rowcolor{gray!10}\multicolumn{12}{c}{\textbf{7B-Scale MLLMs}}\\
\hline
Mantis ~\cite{jiang2024mantis} & 1.52 & 0.00 & 3.31 & 12.18 & 2.08 & 1.00 & 1.01 & 10.02 & 0.00 & 0.85 & 3.20 \\
LLaVA-OV-7B ~\cite{li2024llava1} & 6.06 & 3.19 & 3.43 & 0.18 & 1.04 & 1.08 & 9.09 & 15.43 & 6.93 & 0.85 & 4.73 \\
Minicpm2.6 ~\cite{yao2024minicpm}  & 14.58 & 2.13 & 14.34 & 9.82 & 6.25 & 1.75 & 11.11 & 10.02 & 2.97 & 2.56 & 7.55 \\
mPLUG-Owl3 ~\cite{ye2024mplug} & 18.56 & 6.38 & 34.93 & 8.55 & 7.64 & 2.41 & 7.07 & 22.85 & 9.09 & 5.98 & 12.35 \\
InternVL2-8B ~\cite{chen2024expanding} & 6.92 & 7.45 & 25.49 & 20.73 & 9.72 & 3.49 & 28.28 & 30.26 & 17.82 & 9.40 & 15.96 \\
Qwen2-VL-7B ~\cite{wang2024qwen2} & 27.84 & 38.30 & 19.36 & 20.73 & 11.81 & 25.95 & 23.23 & 58.52 & 48.51 & 11.97 & 28.62 \\
Migician ~\cite{li2025migician} & \textbf{65.15} & 46.81 & \textbf{84.19} & 70.73 & \textbf{60.07} & \textbf{74.31} & 76.77 & 66.53 & 59.41 & 34.19 & 63.82 \\
\rowcolor{blue!8} \textbf{MIRG-RL-7B} & 50.25 & \textbf{53.49} & 83.11 & \textbf{80.75} & 50.35 & 73.67 & \textbf{80.81} & \textbf{71.14} & \textbf{67.84} & \textbf{36.75} & \textbf{64.82} \\
\hline
\end{tabular}
}
\label{tab:mig-bench}
\end{table*}

\section{Experiment and Discussion}

\subsection{Implementation Details}
Following Section~\ref{sec:dataset}, we constructed two datasets from MGrounding-630k~\cite{li2025migician}, covering seven task types (common objects, static differences, reasoning, group grounding, object tracking, referential grounding, and region localization). The cold-start phase utilized 10k samples to enhance reasoning capabilities, while the reinforcement learning phase employed an additional 7k samples containing only final answers.
Based on these constructed datasets, we proceed to train our model through a two-stage strategy.
MIRG-RL was developed on Qwen2-VL-7B~\cite{wang2024qwen2} using 8×80GB GPUs, a global batch size of 128, and learning rates of 2e-5 and 5e-6 for the two phases, respectively. For evaluation on MIG-Bench, we adopted the standard $\text{Acc}_{0.5}$ metric from the referenced expression understanding task~\cite{kazemzadeh2014referitgame}, where a prediction is considered correct if its IoU with the ground truth fact exceeds 0.5.

\subsection{Effectiveness and Superiority of MIRG-RL}
To validate the effectiveness of lightweight and high-quality grounding data with reasoning traces and the feasibility of our image-aware-based reward design, we conducted a series of comparative experiments on MIG-Bench to evaluate the accuracy of inferred answers. As shown in Table~\ref{tab:mig-bench}, our proposed MIRG-RL achieves state-of-the-art results on the MIG-Bench, securing the highest average score of 64.82\% and ranking first on nearly all tasks.

Despite having only 7 billion parameters, MIRG-RL significantly outperforms all 70B-scale models, achieving an average score 25.94\% higher than Qwen2-VL-72B. Among 7B-scale models, MIRG-RL achieves substantial gains over the general baseline, outperforming the base model Qwen2-VL-7B by 35.20\% on average.
Notably, compared to Migician, the state-of-the-art baseline model trained on 630k samples, MIRG-RL achieves a 1\% improvement using only 17k training samples (less than 3\% of Migician's training data). MIRG-RL demonstrates outstanding performance on key tasks, including Object Tracking and Visual Text Reasoning. This remarkable data efficiency, achieving top results, validates the effectiveness of our reasoning-augmented learning strategy.

\subsection{Effects of MIRG-RL on Different Base Models}
\label{base}


As shown in Table~\ref{tab2}, applying our cold-start Trajectory SFT using only the reasoning data from Stage 1 in Figure~\ref{method}(a) significantly improves the performance of Qwen2-VL and Qwen2.5-VL on multi-image tasks. For Qwen2-VL-7B, Trajectory SFT alone increases the average score by 28.70\%, with subsequent reinforcement learning further improving it by 4.31\%. Similarly, Qwen2.5-VL-7B achieves a 5.13\% score improvement after reinforcement learning following Trajectory SFT.

Under identical training configurations, Qwen2-VL consistently outperformed Qwen2.5-VL by 3.05\%, leading us to select Qwen2-VL as the default base model. These results validate that our lightweight Trajectory SFT, combined with reinforcement learning, is both data-efficient and highly effective. By integrating reasoning-enhanced supervision with reinforcement learning, it achieves improvements with minimal training data.

\begin{table}[t] 
\setlength{\tabcolsep}{6pt}
\centering
\caption{COMPARISON OF THE DIFFERENT \textbf{base model} on MIG-Bench. The best result is indicated in \textbf{bold}.}
\vspace{6pt}
\resizebox{\linewidth}{!}{
\begin{tabular}{l|c|c|c}
\toprule
\textbf{Models} &  \textbf{Spontaneous} & \textbf{Referential} & \textbf{AVE} \\
\midrule
Qwen2-VL-7B & 34.29 & 26.79 & 29.04 \\
+ Trajectory SFT  & 55.82 & 58.57 & 57.74 \\
\rowcolor{blue!8}+ Trajectory SFT + RL & 59.47 & 63.15 &  62.05 \\

Qwen2.5-VL-7B & 14.75 & 10.38 & 12.03 \\
+ Trajectory SFT  & 53.84 & 54.31 & 54.17 \\
+ Trajectory SFT + RL & 56.15 & 60.22 & 59.00 \\

\bottomrule
\end{tabular}
}
\label{tab2}
\end{table}

\begin{table}[t] 
\setlength{\tabcolsep}{6pt}
\centering
\caption{COMPARISON OF THE DIFFERENT \textbf{train data} and \textbf{METHODS} ON MIG-Bench. The best result is indicated in \textbf{bold}.}
\vspace{6pt}
\resizebox{\linewidth}{!}{
\begin{tabular}{l|c|c|c}
\toprule
\textbf{Models} &  \textbf{Spontaneous} & \textbf{Referential} & \textbf{AVE} \\
\midrule
Qwen2-VL-72B & 52.46 & 33.05 & 38.88 \\
\midrule
Qwen2-VL-7B & 34.29 & 26.79 & 29.04 \\
+ RL  & 44.47 & 32.23 & 35.90 \\
+ SFT  & 50.93 & 50.98 & 50.96 \\
+ + RL & 56.59 & 54.53 & 55.55 \\
+ Reasoning-only SFT   & 55.82 & 58.57 & 57.74 \\
+ + Reasoning-only RL & 59.47 & 63.15 &  62.05 \\
+ Reasoning and Grounding SFT & 59.54 & 61.05 &  60.60 \\
\rowcolor{blue!8}+ + Reasoning and Grounding RL & 62.28 & 65.90 &  64.82 \\
\bottomrule
\end{tabular}
}
\label{tab3}
\end{table}

\subsection{Ablation Study on Training Data and Method Components}
\label{sec:ablation}

To better understand the contribution of each component in our framework, we conducted a series of ablation studies summarized in Table~\ref{tab3}, which cover both dataset construction and the two-stage training process.

Starting from the original Qwen2-VL-7B baseline, applying reinforcement learning directly to unprocessed data improves performance to 35.90\%. However, as described in Method~\ref{sec:dataset}, not only does it lack multi-image grounding capability, but it also struggles to generate inference trajectories. In contrast, the standard SFT achieves 50.96\%, and SFT+RL reaches 55.55\%. Although these two models exhibit partial multi-image grounding capability, they still struggle to generate inference trajectories. When replacing SFT data with our inference trajectories (Stage 1 in Figure~\ref{method}), the model outperforms standard SFT by 6.78\%, demonstrating substantial gains from reasoning-enhanced supervision.

When upgraded to reasoning trajectories containing image indices and precise bounding boxes, model performance increased by 2.86\%. This indicates that inference trajectories with explicit image and coordinate annotations achieve superior performance. We believe this improvement stems from the model's token-by-token prediction mechanism—having explicit spatial information in the reasoning trajectory increases the probability of predicting correct subsequent tokens. 
When incorporating image-level reward functions into reinforcement learning, model performance continued to show significant gains: a 4.31\% improvement compared to models without this reward. This demonstrates that explicit image annotations significantly improve the model's multi-image grounding performance. 

The final configuration achieved the best result of 64.82\%. These ablations confirm three key findings: (i).\,inference trajectory data is more effective than bbox-caption data alone for SFT.(ii).\,Displaying image information within inference trajectory data is crucial for cross-image linking.
(iii).\,image-aware RL optimization consistently improves inference accuracy.

\section{Conclusions}

In this study, we propose the MIRG-RL multi-image reasoning and grounding framework, which integrates inference-enhanced dataset construction, a two-stage training paradigm, and image-aware rule reinforcement learning. In multi-task experiments, MIRG-RL not only outperforms powerful SFT and RL baseline models but also achieves state-of-the-art performance with significantly less training data. These results demonstrate that combining inference-guided supervision with image-aware reinforcement optimization is an effective and data-efficient strategy for advancing real-world multi-image understanding and grounding tasks.


\bibliographystyle{IEEEbib}

\end{document}